\documentclass{article}

\usepackage{arxiv}

\usepackage{amsmath}
\usepackage[utf8]{inputenc} 
\usepackage[T1]{fontenc}    
\usepackage{hyperref}       
\usepackage{url}            
\usepackage{booktabs}       
\usepackage{amsfonts}       
\usepackage{nicefrac}       
\usepackage{microtype}      
\usepackage{lipsum}
\usepackage{graphicx}
\usepackage{adjustbox}      
\usepackage{multirow}       
\usepackage{siunitx,etoolbox} 
\usepackage{arydshln}       
\usepackage{anyfontsize}
\graphicspath{ {./images/} }

\newcommand{\para}[1]{\noindent\textbf{#1}} 

\usepackage{pifont}
\newcommand{\cmark}{\ding{51}}%

\title{Approximating Families of Sharp Solutions to Fisher's Equation with Physics-Informed Neural Networks}

\author{
  Franz M. Rohrhofer\thanks{corresponding author} \\
  Know-Center GmbH, Research Center for \\
  Data-Driven Business \& Big Data Analytics \\
  Sandgasse 36/4, 8010 Graz, Austria \\
  \texttt{frohrhofer@acm.org} \\
  \And
  Stefan Posch \\
  LEC GmbH, Research Center for \\
  Large Engine Technology \\
  Inffeldgasse 19, 8010 Graz, Austria \\
  \texttt{stefan.posch@lec.tugraz.at} \\
  \And
  Clemens Gößnitzer \\
  LEC GmbH, Research Center for \\
  Large Engine Technology \\
  Inffeldgasse 19, 8010 Graz, Austria \\
  \texttt{clemens.goessnitzer@lec.tugraz.at} \\
  \And
  Bernhard C. Geiger \\
  Know-Center GmbH, Research Center for \\
  Data-Driven Business \& Big Data Analytics\\
  Sandgasse 36/4, 8010 Graz, Austria \\
  \texttt{geiger@ieee.org} \\
}

\begin{document}
\maketitle

\begin{abstract}
This paper employs physics-informed neural networks (PINNs) to solve Fisher's equation, a fundamental reaction-diffusion system with both simplicity and significance. 
The focus is on investigating Fisher's equation under conditions of large reaction rate coefficients, where solutions exhibit steep traveling waves that often present challenges for traditional numerical methods.
To address these challenges, a residual weighting scheme is introduced in the network training to mitigate the difficulties associated with standard PINN approaches. 
Additionally, a specialized network architecture designed to capture traveling wave solutions is explored. The paper also assesses the ability of PINNs to approximate a family of solutions by generalizing across multiple reaction rate coefficients.
The proposed method demonstrates high effectiveness in solving Fisher's equation with large reaction rate coefficients and shows promise for meshfree solutions of generalized reaction-diffusion systems.

\end{abstract}


\section{Introduction}
Reaction-diffusion systems constitute a wide class of mathematical models used in biology, physics, chemistry, ecology, and engineering.
Many dissipative dynamical systems, such as the spread of diseases, dispersion of pollutants, or propagation of flames, can be effectively modeled as reaction-diffusion systems, highlighting their profound significance in scientific research and practical applications.
The mathematical foundation of the underlying differential equation was given in 1937 by Fisher~\cite{fisher1937wave} and Kolmogorov, Petrowskii and Piscounoff~\cite{kolmogorov1937study}, which hence is often referred to as Fisher's, Kolmogorov–Petrovsky–Piskunov, or Fisher–KPP equation. 
Ever since, efforts have been dedicated to the study of Fisher's equation and its well-known travelling wave solutions~\cite{ablowitz1979explicit, wang1988exact, chen2004new, kudryashov2005exact}.
Those solutions are admitted for wave speeds $c\geq2\sqrt{\rho}$, where $\rho$ denotes the reaction rate coefficient that appears as a parameter of the underlying nonlinear parabolic partial differential equation (PDE). 
The reaction rate coefficient determines the steepness of the propagating wave front, which for values $\rho\gg1$ is sharp and commonly referred to as super speed waves~\cite{zhao2003comparison}.
Obtaining an accurate prediction of these traveling wave solutions is a challenging numerical problem, as resolving and tracking the steep wave front demands a fine spatial and temporal resolution.  
In this regard, numerical schemes that are commonly applied on Fisher's equation comprise the sinc collocation method~\cite{al2001numerical}, space derivative method~\cite{gazdag1974numerical}, wavelet Galerkin method~\cite{mittal2006numerical}, and Petrov-Galerkin method~\cite{tang1991numerical}.
Furthermore, also spline-based methods have been applied on Fisher's equation in various B-splines schemes such as cubic~\cite{mittal2013numerical, shukla2016extended, dhiman2018collocation, tamsir2018cubic}, quartic~\cite{csahin2008ab, bacshan2019quartic}, quintic~\cite{sahin2014usage, mittal2016study, dhiman2020numerical}, and exponential~\cite{dag2016exponential, zorsahin2018exponential, tamsir2018numerical}.
These methods generally depend on appropriately selecting the space and time discretization to accommodate the challenging conditions of sharp solutions and steep traveling wave fronts in Fisher's equation.

Recently, physics-informed neural networks (PINNs)~\cite{raissi2019physics} have emerged as a meshfree deep learning method that is applicable to any type of physical system involving differential equations. 
These networks are particularly designed to approximate solutions to differential equations by incorporating data- and physics-based loss functions in the network training.
PINNs have already been successfully applied to various types of physical systems, including fluid-flow~\cite{jin2021nsfnets,eivazi2022physics}, diffusion systems~\cite{shaban2023physics,cai2021physics}, reaction systems~\cite{ji2021stiff,schiassi2022physics} and reaction-diffusion systems~\cite{pan2023high,haitsiukevich2023improved,tarbiyati2023weight}.
One major advantage of PINNs over classical PDE solvers is that they operate in a mesh- and discretization-free manner.
This is achieved by the continuous form of the neural network function as a global ansatz, and the minimization of a physics loss function which uses collocation points to minimize residuals on the differential equation. 
The collocation points can be randomly sampled from the computational domain, which gives great flexibility in the number and position of residuals that should be minimized.
Training PINNs, however, is not straightforward and demands a carefully chosen optimization procedure in order to converge to the right solution of the physics loss function~\cite{krishnapriyan2021characterizing,liu2024discontinuity,mao2023physics,de2024physics}.
Discontinuous and sharp solution functions often seem to cause issues for the optimization why PINNs have been primarily applied to the Fisher's equation with small reaction rate coefficients and smooth traveling wave fronts~\cite{haitsiukevich2023improved,tarbiyati2023weight}. 

In this paper, PINNs are employed to approximate solutions to Fisher's equation with large reaction rate coefficients $\rho$, particularly exploring values ranging from $10^2$ to $10^4$.
This range is commonly studied in the literature with $\rho=10^4$ representing a strong reaction problem, where the solution evolves into a sharp, shock-like wave~\cite{zhao2003comparison,gazdag1974numerical,li1998stability,qiu1998numerical}.
To address the optimization challenges faced by standard PINNs in approximating sharp wavefronts, a novel residual weighting method is introduced. This method effectively reduces the influence of the physics residuals near sharp transitions, stabilizing the optimization process for cases with large reaction rate coefficients $\rho\gg1$.
Furthermore, the performance of a specific network architecture designed to adapt to the shape of traveling wave fronts is evaluated. In the final experiment, the generalization capability of PINNs is explored by training a single PINN to approximate a family of traveling wave solutions. 
This is achieved by incorporating the reaction rate coefficient as an additional input, enabling applications for solving reaction-diffusion systems across a continuous parameterization domain.

\section{Reaction-Diffusion Problem}~\label{sec:reaction_diffusion}
The most general form of a reaction-diffusion system, referred to as the Kolmogorov-Petrovsky-Piskunov (KPP) equation~\cite{kolmogorov1937study}, is given by
\begin{equation}
	\frac{\partial u}{\partial t} - \mu \frac{\partial^2 u}{\partial x^2} = F(u;\rho),
	\label{eq:reaction_diffusion}
\end{equation}
where $u$ refers to the concentration of a chemical substance, $\mu$ describes the diffusion coefficient, and $F(u;\rho)$ is the reaction term parameterized by the reaction rate coefficient $\rho$.
The solution function $u$, as well as the reaction term $F(u)$, are continuous nonlinear functions satisfying
\begin{equation}\label{eq:reaction_props}
	\begin{aligned}
		&F(0)=F(1)=0; \\
		&F(u)>0, \quad\mathrm{for}\quad0<u<1; \\
            &F'(0)=a>0; \quad F'(u)<a \quad\mathrm{for}\quad 0<u\leq1.
	\end{aligned}
\end{equation}
This model describes the interaction between diffusion transport and reaction mechanism, respectively given by the left-hand and right-hand side of Eq.~\eqref{eq:reaction_diffusion}.

\subsection{Fisher's Equations}~\label{sec:fishers_equation}
A fundamental and well-studied reaction term was given in~\cite{fisher1937wave} and is defined by
\begin{equation}
	F(u;\rho)=\rho u(1-u).
	\label{eq:reaction_mechanism}
\end{equation}
Inserting this term into Eq.~\eqref{eq:reaction_diffusion} yields the so-called Fisher's equation:
\begin{equation}\label{eq:fishers_equation}
	\frac{\partial u}{\partial t} - \frac{\partial^2 u}{\partial x^2} = \rho u(1-u),
\end{equation}
where the dependency on $\mu$ has been removed through nondimensionalization using $x\leftarrow x/\sqrt{\mu}$.
This equation has two natural equilibrium states (or fixed points) that are $u^*=0$ and $u^*=1$, where $u^*=0$ is an unstable and $u^*=1$ is a stable fixed point.

\subsection{Traveling Wave Solutions}
Fisher's equation admits traveling wave solutions for wave speeds $c\geq2\sqrt{\rho}$.
These solutions are given in the general form
\begin{equation}\label{eq:wave_equation}
	u(x,t)=u(x\pm ct)\equiv u(z),
\end{equation} 
and satisfy
\begin{equation}
	\lim_{z\to-\infty}u(z)=0, \quad \lim_{z\to\infty}u(z)=1.
\end{equation}

\para{Steep Traveling Waves.}
While the dependency of the reaction rate coefficient $\rho$ in Eq.~\eqref{eq:fishers_equation} can be eliminated through further nondimensionalization, studying traveling waves often retains the dependency on $\rho$. 
This choice enables the nonlinear reaction term in Eq.~\eqref{eq:fishers_equation} to be arbitrarily larger than the diffusion term, thereby adjusting the wave speed and, consequently, the steepness of the solution function for a fixed computational domain~\cite{zhao2003comparison, olmos2006pseudospectral}.

\para{Analytical Solution.}
The traveling wave solution with a constant wave speed of $c=5\sqrt{\rho/6}$ often serves as test case for numerical studies since it has been analytically derived in~\cite{ablowitz1979explicit}.
The solution function for this specific wave speed reads
\begin{equation}
	u(x,t;\rho)=\left[1+\exp\left(\sqrt{\frac{\rho}{6}}x-\frac{5\rho}{6}t \right) \right]^{-2}.
	\label{eq:fishers_solution}
\end{equation}

\para{Challenges.} 
Solving Fisher's equation with large reaction rates pose a challenging numerical problem, requiring careful resolution and tracking of the traveling wave front due to its steep nature.
For $\rho\gg1$, such solutions are sometimes referred to as super speed waves~\cite{zhao2003comparison}, and of particular interest in measuring the performance of numerical discretization schemes~\cite{gazdag1974numerical,li1998stability,qiu1998numerical}.

\begin{figure}
    \centering
    \includegraphics[width=1\textwidth]{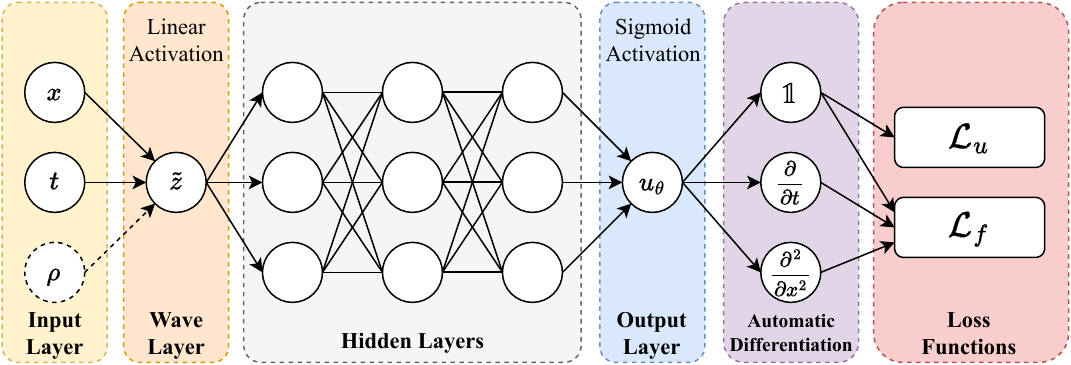}
    \caption{Schematic drawing of the \textit{wave}-PINN that either approximates $u(x,t)$, or $u(x,t;\rho)$ by taking $\rho$ as an additional input variable. The additional wave layer (see Fig.~\ref{fig:wave_layer}) constrains the network function to $u_\theta(\tilde z)$, thus ensuring the form of traveling wave solutions. The \textit{standard} architecture does not use the wave layer. The \textit{generalizing} architecture uses $\rho$ as an additional input.}    
    \label{fig:wave_PINN}
\end{figure}

\section{Physics-Informed Neural Networks}\label{sec:fishers_PINNs}
A comprehensive description of PINNs can be found in~\cite{raissi2019physics}.
For the sake of clarity, this section provides their fundamentals and discusses particular parts of the PINN framework needed in order to understand the modification being made in this work.
In essence, PINNs represent a particular class of neural networks utilized for approximating solutions to differential equations by incorporating data- and physics-based soft constraints (or loss functions) during network training.
The inputs to the network structure include the spatial and temporal coordinates of the physical system, though as it will be discussed shortly, they may not be limited to those variables.
The network's output corresponds to the targeted physical quantity; in this work, the concentration $u$ governed by Eq.~\eqref{eq:reaction_diffusion}. 
By selecting smooth activation functions for neurons in the hidden layers, e.g., hyperbolic tangent (tanh) or Sigmoid linear unit (swish), the network function is continuous over the (spatial-temporal) input domain.
Accumulating all weights and biases of the network into a single network weight variable $\theta$, the network function is referred to as $u_\theta$.

\subsection{Architectures}
In this work, two distinct network architectures are tested: a \textit{standard} architecture and a \textit{wave} architecture. 
Additionally, the study comprises a network architecture designed for a discrete-$\rho$ approximation and a \textit{generalizing} architecture for a continuous-$\rho$ approximation.
The \textit{generalizing} architecture considers the reaction rate coefficient as an additional input, thus approximates the solution for multiple values of $\rho$.

\begin{figure}
    \centering
    \includegraphics[width=0.75\textwidth]{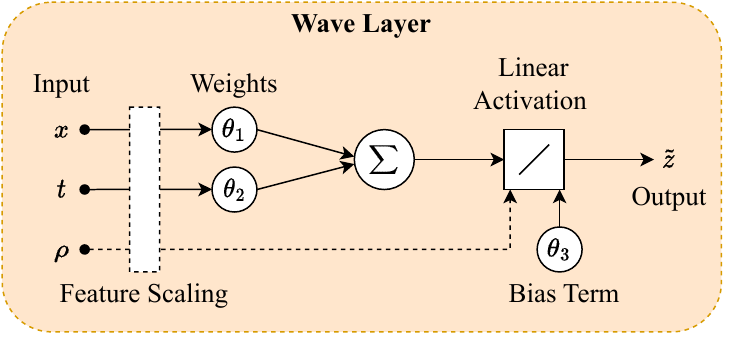}
    \caption{Schematic structure of the wave layer, which compresses the input variables into the form $\tilde z(x,t) = \theta_1 x + \theta_2 t + \theta_3$ for the single-$\rho$ approximation, or $\tilde z(x,t;\rho) = \theta_1 \rho_1 x + \theta_2 \rho_2 t + \theta_3$ when using the \textit{generalizing} architecture. Specific transformations in the feature scaling step are used to obtain $\rho_1$ and $\rho_2$, and to ease the adjustment of $\theta_1$, $\theta_2$, and $\theta_3$.}    
    \label{fig:wave_layer}
\end{figure}

\para{Standard and Wave Architectures.}
In the \textit{standard} architecture, hidden layers directly follow the input layer, aiming to approximate the solution function in the general spatial-temporal form $u(x,t)$. 
Conversely, in the \textit{wave} architecture, an additional wave layer is introduced directly after the input layer and before the hidden layers, see Fig.~\ref{fig:wave_PINN}.
The wave layer introduces a latent variable representation, as depicted in Fig.~\ref{fig:wave_layer}.
The layer's output is $\tilde z=\theta_1x+\theta_2t+\theta_3$, 
where $\theta_1,\theta_2$, and $\theta_3$ represent trainable parameters.
Consequently, the network function is compelled to adopt the traveling wave form $u(\tilde z)$ as described in Eq.~\eqref{eq:wave_equation}.
While for the discrete-$\rho$ approximation the wave layer is a standard linear layer, additional modifications are introduced for the \textit{generalizing} architecture, as it will be discussed shortly.
It is essential to note that the additional bias term $\theta_3$ plays a crucial role, as data preprocessing steps, such as feature scaling, can significantly influence the effectiveness of a simpler representation with fewer parameters.

\para{Generalization.}
This work also evaluates the capability of a single PINN instance in approximating solutions to Eq.~\eqref{eq:fishers_equation} for multiple and continuous values of $\rho$. 
Referred to as the \textit{generalizing} architecture, this is achieved by treating the reaction rate coefficient $\rho$ as an additional input parameter.
The network architecture thus represents an approximation to a family of solutions in the form $u(x,t;\rho)$.

This additional input is either directly connected with the hidden layers (\textit{standard} architecture) or processed in the wave layer (\textit{wave} architecture). 
In the wave layer, the reaction rate coefficient is coupled with the spatio-temporal input variables through $\tilde z =\theta_1 \rho_1 x+\theta_2 \rho_2 t+\theta_3$, as captured by Fig.~\ref{fig:wave_layer}.
Here, $\rho_1$ and $\rho_2$ are obtained through specific transformation used in the feature scaling which will be further discussed in Section~\ref{sec:experimental_setup_fisher}.

Treating $\rho$ as an additional input variable enables a straightforward extension of the application of sampling schemes for collocation points.

\subsection{Loss Functions}
The data loss function which encodes initial conditions (IC) and boundary conditions (BC), or incorporates additional training data, is given by
\begin{equation}\label{eq:data_loss_function}
	\mathcal{L}_{u}(\theta)=\frac{1}{N}\sum_{i=1}^{N}|u^{(i)}-u^{(i)}_\theta|^2,
\end{equation}
where $u^{(i)}_\theta$ denotes the network's output evaluated at $N$ data points from the labeled dataset $\left\{\left(x^{(i)}, t^{(i)};\rho^{(i)}\right), u^{(i)}\right\}_{i=1}^N$.

The physics loss function is evaluated on $N_\mathrm{col}$ collocation points from the unlabeled dataset $\left\{\left(x^{(i)}, t^{(i)};\rho^{(i)}\right)\right\}_{i=1}^{N_\mathrm{col}}$, and
reads
\begin{equation}
	\mathcal{L}_f(\theta)=\frac{1}{N_\mathrm{col}}\sum_{i=1}^{N_\mathrm{col}}\left|f^{(i)}_\theta\right|^2,
\end{equation}
where the physics residuals $f_\theta$ encoding Eq.~\eqref{eq:fishers_equation} are given by

\begin{equation}\label{eq:physics_residuals}
	f_\theta(x,t;\rho) = \frac{\partial u_\theta}{\partial t} - \frac{\partial^2 u_\theta}{\partial x^2} - \rho u_\theta(1-u_\theta).
\end{equation}
Both loss function are finally combined by the total loss function that reads
\begin{equation}\label{eq:total_loss_fisher}
	\mathcal{L}(\theta) = \mathcal{L}_u(\theta) + \mathcal{L}_f(\theta).
\end{equation}

\subsection{Residuals Weighting}\label{sec:residual_weighting}
It is well known that the standard PINN framework has issues with sharp and discontinuous solutions~\cite{krishnapriyan2021characterizing,liu2024discontinuity,mao2023physics,de2024physics}.
As reported in~\cite{liu2024discontinuity}, this phenomenon can be attributed to a paradoxical state at transition points, where the steep trend of the solution leads to nearly explosive behavior in the network derivatives and physics residuals in Eq.~\eqref{eq:physics_residuals}.
Minimizing these residuals consequently results in reducing the network derivatives toward a smooth, potentially incorrect, solution that fails to capture the sharpness and steepness of the traveling wave. 

\para{Reaction Term Weighting.}
Weighting the residuals has been already proposed as a remedy to cope with challenging solution functions \cite{liu2024discontinuity,mao2023physics}.
As a novel way of weighting the residuals for Fisher's equation, and reaction-diffusion systems in general, a weighting scheme according to the strength of the reaction term $F$ is proposed.
The motivation behind this lies in the fact that the sharpness of traveling waves is largely influenced by the reaction dynamics.
In view of that, weighting each physics residual is performed according to
\begin{equation}
	\omega=\frac{1}{\lambda|F(u_\theta;\rho)| + 1} .
	\label{eq:collocation_weighting}
\end{equation}
Here, $\lambda$ is a hyperparameter that can be used to adjust the strength of the weighting, and trimmed to the particular problem given at hand.
By weighting residuals according to Eq.~\eqref{eq:collocation_weighting}, residuals in smooth regions with predominant diffusion consequently get a weight of $\omega\to1$.
However, weights of residuals in sharp regions are lowered, i.e. $\omega\to0$, which effectively suppresses the effect of large gradients and physics loss residuals.

Using this weighting scheme in the definition of the physics loss function yields
\begin{equation}
	\mathcal{L}_f(\theta)=\frac{1}{N_\mathrm{col}}\sum_{i=1}^{N_\mathrm{col}}\left|\omega^{(i)} f^{(i)}_\theta \right|^2,
	\label{eq:loss_physics_weighted}
\end{equation}
where a value of $\lambda=0$ corresponds to the unweighted form of the physics loss function as used by the conventional PINN framework. 

\begin{table}
    \centering
    \caption[Models used in the study of Fisher's equation]{The four models employed in the study. These can be distinguished based on whether the physics loss function and/or the wave layer architecture are applied.} 
    \label{tab:models_fishers}
    \renewcommand{\arraystretch}{1.5}
    \begin{tabular}{@{}r|cc@{}}
        \toprule
        Model         & Physics Loss & Wave Layer \\ 
        \midrule
        \textit{standard}-ANN  &        &        \\
        \textit{wave}-ANN      &        & \cmark \\
        \textit{standard}-PINN & \cmark &        \\
        \textit{wave}-PINN     & \cmark & \cmark \\ 
        \bottomrule
    \end{tabular}
\end{table}

\section{Experimental Setup}\label{sec:experimental_setup_fisher}

For the experiments in the next section, four distinct models are employed that are listed in Table~\ref{tab:models_fishers}. All code has been made available and can be accessed at: \url{https://github.com/frohrhofer/PINN_fishers}

\para{Data-driven ANNs.}
The first two models are data-driven ANNs which serve as a reference for our study.
These models learn the solution function directly from labeled data, sampled from the analytical solution~\eqref{eq:fishers_solution} and fed through the data loss function~\eqref{eq:data_loss_function}.
The data is sampled anew at each training iteration, allowing access to an infinite number of training data points which is used to prevent overfitting. 
Notably, these models do not consider the physics loss function and hence serve as a reference for assessing how accurately the solution function can be learned purely from data.
This setting is tested using both the \textit{standard} architecture and the \textit{wave} architecture, resulting in the two considered data-driven models.

\para{PINNs.}
The other two models are PINN instances that apply the methodology discussed in Section~\ref{sec:fishers_PINNs}. 
These models infer the solution functions by minimizing the loss function given in Eq.~\eqref{eq:total_loss_fisher}, where the data loss function is used to enforce the initial and boundary conditions (IC and BC).
This setup is investigated using both the \textit{standard} architecture and the \textit{wave} architecture. 
As part of initial experiments, different settings for the residual weighting scheme are evaluated, by considering $\lambda \in \{0, 0.1, 1, 10\}$ as defined in Eq.~\eqref{eq:collocation_weighting}. 
Notably, $\lambda=0$ corresponds to unweighted optimization, which aligns with conventional PINN optimization.

\para{Computational Domain.}
Fisher's equation is studied with reaction rate coefficients in the range $\rho\in[10^2,10^4]$.
To sufficiently resolve the traveling wave front for any value of $\rho$ within the computational domain, the spatial domain is defined as $x \in [-5, 5]$, and the temporal domain as $t \in [0, 0.004]$.
These settings are commonly adopted in numerical studies~\cite{olmos2006pseudospectral,li1998stability,mittal2010efficient}.
Furthermore, the system is examined with a diffusion coefficient $\mu=10$ in 
 Eq.~\eqref{eq:reaction_diffusion}.
This choice ensures that both terms in the loss function of Eq.~\eqref{eq:total_loss_fisher} remain within a similar range, thereby eliminating the need for an additional loss weighting~\cite{rohrhofer2023apparent}. 
Importantly, by applying the transformation $\hat x \leftarrow x/\sqrt{\mu}$, the setup remains consistent with those used in previous numerical studies, where the analytical solution is given by Eq.~\eqref{eq:fishers_solution}.

\para{Feature Scaling.}
Scaling the features in this study has proven to be essential for training any of the models within the chosen computational domain. 
Two distinct feature scaling strategies are employed, contingent on whether the \textit{standard} or \textit{wave} architecture is utilized.
In the \textit{standard} architecture, all input variables undergo scaling to the unit range $[0,1]$.
Conversely, in the \textit{wave} architecture, a different scaling strategy is implemented to ensure reduced sensitivity of the $\theta_j$ weights in the wave layer (refer to Fig.~\ref{fig:wave_layer}). 
This is achieved by leaving the spatial variable $x$ unscaled, while the temporal variable is scaled to the unit range $[0,1]$.
The reaction rate input, however, undergoes a specific transformation given by $\rho_1 \leftarrow \sqrt{\rho}$, while $\rho_2=\rho$.
Notably, this transformation is motivated by the commonly performed nondimensionalization steps with $\hat x\leftarrow \sqrt{\rho}x$ and $\hat t\leftarrow \rho t$, which are applied to reaction-diffusion systems in the form of Eq.~\eqref{eq:fishers_equation}.

\para{Data Settings.}
Due to the finite computational domain, IC and BCs are imposed by randomly sampling data from the analytical solution at the boundary of the computational domain.
These data points are provided to the PINN through the data loss function.
The collocation points are randomly sampled inside the computational domain using the Latin hypercube sampling scheme.
Datasets for both IC/BCs and collocation data points are sampled anew at each training epoch with a default dataset size of $N = 1024$, respectively.
The data for the data-driven ANNs is also sampled anew at each epoch with a size of $N=1024$, to ensure that these models are trained at the same pace as the PINN models.

\para{Network Settings.}
For the network architecture applied for discrete values of $\rho$, two hidden layers and 20 neurons per hidden layers are taken.
The \textit{generalizing} architecture, however, is applied with three hidden layers, and the same amount of neurons, to accommodate the increased complexity introduced by the additional dimension.
The hyperbolic tangent (tanh) is used as hidden layer activation and the Sigmoid activation as output layer activation, to ensure that any prediction lies in the bounded interval $u_\theta\in[0,1]$.

In all experiments, the Glorot weight initialization scheme is applied and Adam as the gradient-based optimization algorithm for minimizing the loss function given by Eq.~\eqref{eq:total_loss_fisher}. 
Optimization is carried out for $50k$ training epochs in the discrete-$\rho$ approximation and for $100k$ epochs in the continuous-$\rho$ approximation.
For all subsequent experiments, the most effective optimization settings were found to be an initial learning rate of $0.001$ with an exponential decay rate of $0.95$ every $1,000$ epochs.
Consequently, these settings are consistently applied throughout the experiments presented in this manuscript.

\para{Performance Measure.} The PINN's performance is evaluated using the $L_2$-error which is defined as 
\begin{equation}
	L_2=\left(\frac{1}{N}\sum_{i=1}^N\left|u^{(i)}-u_\theta^{(i)} \right|^2 \right)^{1/2},
\end{equation}
where $u^{(i)}$ represents the analytical solution from Eq.~\eqref{eq:fishers_solution}, and $u_\theta^{(i)}$ denotes the network's prediction for the $i$-th data point in a test set consisting of $N$ data points.
The test set is constructed from a uniform $100$x$100$ grid over the spatio-temporal domain.

\section{Results}\label{sec:fishers_results}

\subsection{Discrete-$\rho$ Approximation}\label{sec:discrete_rho}
For the discrete-$\rho$ experiment, the model performance is evaluated by approximating the solution function for three individual reaction rate coefficients $\rho\in\{10^2,10^3,10^4\}$, training a separate model for each coefficient.
Notably, with an increasing $\rho$ the solution function shifts to a sharp traveling wave since the computational domain is kept equal for all cases. 
Ten differently initialized instances are trained for each of the four models and each of the three coefficients.

\begin{table}
    \centering
    \caption{The mean (and standard deviations) of the $L_2$-error for the four tested models and different values for $\rho$. All numbers should be multiplied by $10^{-4}$. \textbf{Bold}: Best results.} 
    \label{tab:discrete_rho}
    \begin{tabular}{rl|ccc}
        \toprule
        & & \multicolumn{3}{c}{\textbf{Reaction rate coefficient}, $\rho$} \\
        \textbf{Model} & $\lambda$ & \multicolumn{1}{c}{$10^2$} & \multicolumn{1}{c}{$10^3$} & \multicolumn{1}{c}{$10^4$} \\
        \midrule
        \textit{standard}-ANN & - & 0.82 (0.17) & 0.74 (0.09) & 1.72 (0.24) \\
        \textit{wave}-ANN & - & 0.64 (0.11) & 0.64 (0.08) & 1.35 (0.45) \\ \hline
        \multirow{4}{*}{\textit{standard}-PINN} & 0 & 170 (31) & 1799 (1972) & 6375 (3465) \\ 
        & 0.1 & 69.3 (16.2) & 103 (20) & 742 (519) \\ 
        &  1 & 8.33 (1.66) & 19.5 (4.4) & 324 (171) \\
        & 10 & 0.91 (0.22) & 8.66 (1.23) & 458 (456) \\ \hline
        \multirow{4}{*}{\textit{wave}-PINN} & 0 & 82.9 (120) & 742 (636) & 6670 (3051) \\ 
        & 0.1 & 5.58 (7.11) & 164 (456) & 1400 (2407) \\ 
        & 1 & \textbf{0.14 (0.09)} & \textbf{0.06 (0.03)} & \textbf{0.31 (0.28)}  \\
        & 10 & 0.15 (0.09) & 0.16 (0.08) & 0.71 (0.55)  \\
        \bottomrule                   
    \end{tabular}
\end{table}

\para{Result Table.}
The results to this experiment can be found in Table~\ref{tab:discrete_rho}, which displays the mean (and standard deviations) of the $L_2$-error for each tested model and setting. 
For the sake of clarity, the best result has been highlighted for each value of the reaction rate coefficient $\rho$.

\para{Best Model.}
First, attention is directed to the highlighted (bold) numbers in the table, which indicate the model achieving the lowest $L_2$-error for each $\rho$. 
From the table it is apparent that the wave-PINN with $\lambda=1$ consistently achieves the lowest error across all tested reaction rates, even outperforming both data-driven ANNs.
This result is particularly intriguing, given that the data-driven ANNs effectively learned the solution from the data without overfitting, demonstrating high accuracy for the given network size.

\para{Standard vs. Wave.}
By comparing models using the \textit{standard} and \textit{wave} architecture, it can be generally found that the \textit{wave} architecture performs better, with a substantial difference within the PINN models.
The results clearly show that for PINN setting with $\lambda\geq1$, the \textit{wave} architecture achieves significantly lower prediction errors than the \textit{standard} architecture. 

\para{Residual Weighting.}
A clear trend emerges regarding the effectiveness of the proposed residual weighting scheme.
The conventional, unweighted PINN optimization ($\lambda=0$) exhibits significant performance issues, particularly with large $L_2$-errors for higher $\rho$ values, as apparent from Table~\ref{tab:discrete_rho}. 
The results further indicate that applying any form of residual weighting ($\lambda>0$) mitigates these issues and improves the performance. 
The impact of residual weighting becomes particularly pronounced for $\lambda\geq1$, with $\lambda=1$ consistently providing optimal results for the \textit{wave}-PINN.
In contrast, the optimal choice of $\lambda$ is less straightforward for the \textit{standard}-PINN.
For $\rho=10^2$ and $\rho=10^3$, a weighting factor of $\lambda=10$ delivers better performance, whereas for $\rho=10^4$, the best results are achieved with $\lambda=1$.

\begin{figure}
    \centering
    \includegraphics[width=\textwidth]{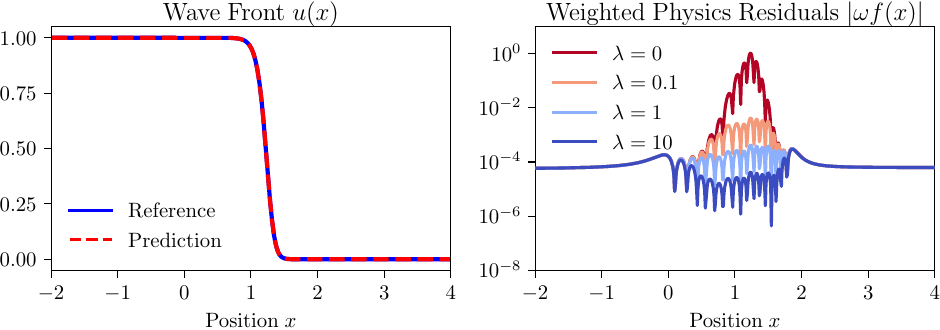}
    \caption{Wave front for $\rho=10^4$ and $t=0.002$. \textit{Left}: Reference solution and prediction of the \textit{wave}-PINN with the fully trained model using $\lambda=1$. \textit{Right}: Weighted physics residuals for different values of $\lambda$ (see Eq. \eqref{eq:collocation_weighting}), assessed for the predicted wave front in the left plot. Unweighted physics residuals ($\lambda=0$) exhibit orders of magnitude larger values at transition points on the wave front compared to those outside the wave front. }
	\label{fig:fisher_wave_front}
\end{figure}

To further investigate the effectiveness of the residual weighting, the wave front and weighted physics residuals for $\rho=10^4$ is illustrated in Fig.~\ref{fig:fisher_wave_front}.
In the figure, the left plot presents the true and predicted wave front for the time $t=0.002$.
The prediction is based on the fully trained \textit{wave}-PINN that utilizes $\lambda=1$.
To examine the effects of the residual weighting scheme, the right plot displays the weighted physics residuals evaluated for this predicted solution using $\lambda\in\{0,0.1,1,10\}$.
Apparent from the figure is that for the unweighted case ($\lambda=0$), physics residuals along the steep wave front take values orders of magnitude larger than residuals outside the wave front and in the smooth region.
This coincides with findings discussed in~\cite{liu2024discontinuity},  where it is expected that further optimization using $\lambda=0$ will drive away the accurately learned solution to a smoother, potentially incorrect wave front.
It can be further observed that applying the residual weighting scheme with values $\lambda>0$ effectively reduces the strength of residuals along the wave front, which, in particular for $\lambda=1$, results in a more balanced allocation of physics residuals along the wave front. 
Notably, $\lambda=10$ further reduces the strength of the residuals along the wave front, which can take even smaller values than residuals in the smooth regions. 
This over-adjustment could explain why in Table~\ref{tab:discrete_rho} it is found that the \textit{wave}-PINN with $\lambda=1$ achieves better results than with $\lambda=10$.

\begin{figure}
    \centering
    \includegraphics[width=\textwidth]{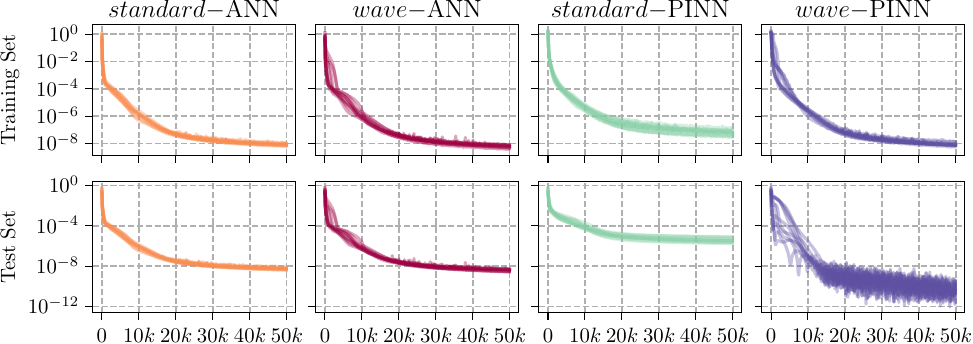}
    \caption{Learning curves for the four tested models for $\rho=10^3$, using ten uniquely initialized instances per model. For the PINN models, results are shown for $\lambda=1$ only. (\textit{Top row}) Training set errors, evaluated using the total loss function in Eq.~\eqref{eq:total_loss_fisher} (\textit{Bottom row}) Test set errors, evaluated by calculating the data loss from Eq.\eqref{eq:data_loss_function} on randomly sampled test points within the computational domain.}
    \label{fig:learning_curves}
\end{figure}

\para{Learning Curves.}
To further analyze the error reduction process during training, the learning curves for $\rho=10^3$ are presented in Fig.~\ref{fig:learning_curves}.
The plot shows the training and test set errors for ten uniquely initialized instances per model, with results for the PINN models displayed for $\lambda=1$ only.
While all models demonstrate stable training behavior as indicated by the training set errors, noticeable variation is observed in the test set errors.
Both data-driven models exhibit similar performance, with the \textit{wave}-ANN achieving slightly lower test set errors.
However, the optimization of the \textit{standard}-PINN appears to be hindered by the issues previously mentioned: although the training set error decreases, the test set errors remain relatively high, indicating that the physics loss function does not adequately resolve the traveling wave front.
In contrast, for the \textit{wave}-PINN, the test set errors exhibit significant diversity in the early training stages but consistently surpass the errors of the data-driven models after approximately $10k$ epochs, eventually settling at the lowest test set error levels.
These findings are qualitatively consistent with the results presented in Table~\ref{tab:discrete_rho}.

\begin{table}
    \centering
    \caption{The mean (and standard deviations) of the $L_2$-error for various model settings and different values for $\rho$. The baseline model represents the \textit{wave}-PINN ($\lambda=1$) with a 2x20 architecture, \textit{tanh} activation function and $N_{col}=1024$ collocation points. All numbers should be multiplied by $10^{-4}$.} 
    \label{tab:ablation_study}
    \begin{tabular}{rl|ccc}
        \toprule
        & & \multicolumn{3}{c}{\textbf{Reaction rate coefficient}, $\rho$} \\
        \textbf{Settings} & & \multicolumn{1}{c}{$10^2$} & \multicolumn{1}{c}{$10^3$} & \multicolumn{1}{c}{$10^4$} \\
        \midrule
        Baseline &  & 0.14 (0.09) & 0.06 (0.03) & 0.31 (0.28) \\ \hline
        \multirow{2}{*}{Layers and Neurons} & 2x10 & 0.10 (0.06) & 0.08 (0.09) & 0.74 (0.60)  \\ 
        & 2x30 & 0.14 (0.10) & 0.09 (0.04) & 0.28 (0.21)  \\ \hline
        \multirow{3}{*}{Activation} & swish & 0.05 (0.04) & 0.08 (0.09) & 3.06 (1.88)  \\ 
        & sigmoid & 0.08 (0.04) & 0.14 (0.07) & 0.85 (0.77)  \\
        & sine & 0.18 (0.15) & 5.13 (8.54) & 203 (20)  \\ \hline
        \multirow{2}{*}{$N_{col}$} & 512 & 0.20 (0.11) & 0.11 (0.08) & 0.42 (0.38) \\ 
        & 2048 & 0.18 (0.10) & 0.07 (0.03) & 0.27 (0.12)  \\
        \bottomrule                   
    \end{tabular}
\end{table}

\para{Ablation Study.}
Finally, an ablation study is conducted to assess the influence of various model and optimization settings. 
The \textit{wave}-PINN with $\lambda=1$, as reported in Table~\ref{tab:discrete_rho}, serves as the baseline model, with individual changes made to either the model architecture (layers, neurons, activation functions) or the optimization settings (number of collocation points, $N_{col}$).
For each configuration, ten uniquely initialized instances are trained and evaluated. The results are summarized in Table~\ref{tab:ablation_study}.
The table reveals that for $\rho=10^2$, the model's performance is relatively insensitive to the modifications. 
However, for larger reaction rate coefficients, the model performance shows a noticeable trend with respect to network architecture and the number of collocation points. 
Specifically, larger architectures and an increased number of collocation points tend to reduce the $L_2$-error.
Regarding activation functions, the baseline model using the tanh activation function performs best for larger reaction rate coefficients, indicated by the lowest $L_2$-error. 
In contrast, the use of the sine activation function leads to significant performance degradation for larger reaction rate coefficients.

\subsection{Continuous-$\rho$ Approximation}
In this section, the \textit{generalizing} network architecture is tested by incorporating $\rho$ as an additional input.
Subsequently, the models approximate the solution $u(x,t;\rho)$ to Eq.~\eqref{eq:fishers_equation} across the continuous parameterization domain $\rho\in[\rho_\mathrm{min},\rho_\mathrm{max}]$.

\para{Assessing Interpolation Capability.}
A specific training strategy is applied to assess the data-driven ANN and PINN interpolation capability. 
In this approach, each model is trained using labeled data sampled from the analytical solution for two specific reaction rate coefficients, denoted as $\rho_\mathrm{min}$ and $\rho_\mathrm{max}$. 
Consequently, the interpolation task is to find good approximations within the continuous range $\rho\in[\rho_\mathrm{min},\rho_\mathrm{max}]$.
These coefficients define the boundaries of the computational domain in the $\rho$-dimension, and determine the range in which collocation points are uniformly sampled for the PINN models using Latin hypercube sampling. 
For the experiment, three distinct test cases are considered using different $\rho$-domains with $\rho_\mathrm{min}\in\{10^2, 10^3\}$ and $\rho_\mathrm{max}\in\{10^3, 10^4\}$.
Again, ten differently initialized instances are trained for each model and setting.

\para{Result Table.}
The results to this experiment can be found in Table~\ref{tab:continuous_rho}, which displays the mean (and standard deviations) of the $L_2$-error for each tested model and setting. 
Notably, the $L_2$-error was assessed on test sets, uniformly sampled across the domain $\rho\in(\rho_\mathrm{min},\rho_\mathrm{max})$. 
For the sake of clarity, the best result for each specific setting within the $\rho$-domain has been highlighted.

\para{Best Model.}
Consistent with the results from the previous experiment, the \textit{wave}-PINN achieves the lowest prediction error in all three cases. 
Interestingly, the \textit{wave}-ANN also performs astonishingly well, with its errors marginally worse.
This model, trained only on the two reaction rate coefficients $\rho_\mathrm{min}$ and $\rho_\mathrm{max}$, proves capable of achieving accurate interpolation within this range.
Evidently, the incorporation of the wave layer provides a highly effective constraint that aids in finding solutions over a continuous parameterization.

\begin{table}
    \centering
    \caption[Continuous-$\rho$ approximation for Fisher's equation]{The mean (and standard deviations) of the $L_2$-error for the four tested models over a specific domain  $\rho\in(\rho_\mathrm{min},\rho_\mathrm{max})$. All numbers should be multiplied by $10^{-4}$. \textbf{Bold}: Best results.} 
    \label{tab:continuous_rho}
    \begin{tabular}{@{}rcl|llll@{}}
        \toprule
        $\rho_\mathrm{min}$ & & $\rho_\mathrm{max}$ & \textit{standard}-ANN & \textit{wave}-ANN & \textit{standard}-PINN & \textit{wave}-PINN \\ 
        \midrule
        $10^2$ & - & $10^3$ & 110 (51) & 1.83 (1.54) & 5.16 (0.66) & \textbf{1.46 (1.79)} \\
        $10^3$ & - & $10^4$ & 559 (101) & 3.21 (2.96) & 173 (23) & \textbf{1.83 (1.27)} \\
        $10^2$ & - & $10^4$ & 1354 (120) & 2.15 (2.23) & 363 (45) & \textbf{1.30 (1.69)} \\ 
        \bottomrule
    \end{tabular}
\end{table}

\begin{figure}
    \centering
    \includegraphics[width=\textwidth]{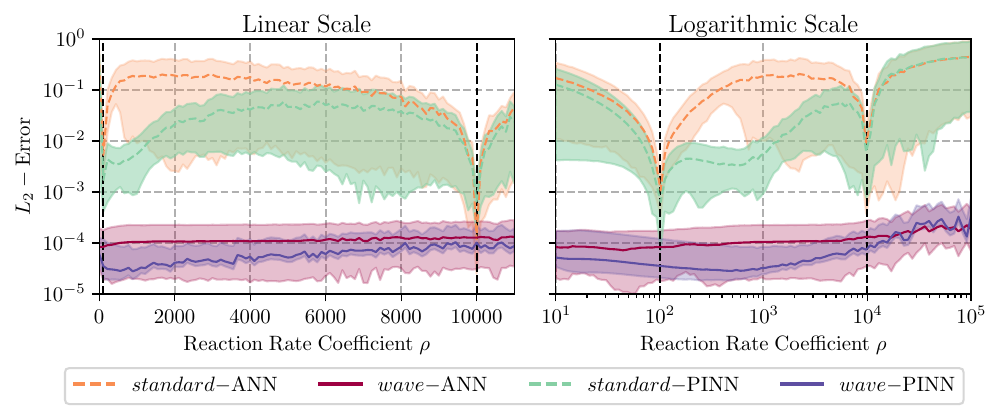}
    \caption{The median (and 25\%- and 75\%-quantiles) of the $L_2$-error for each tested model on the continuous domain $\rho\in[100, 10.000]$. Labeled training data was used at $\rho=10^2$ and $\rho=10^4$, indicated by the dashed black lines.}
    \label{fig:continuous_rho_interpolation}
\end{figure}

\begin{figure}
    \centering
    \includegraphics[width=\textwidth]{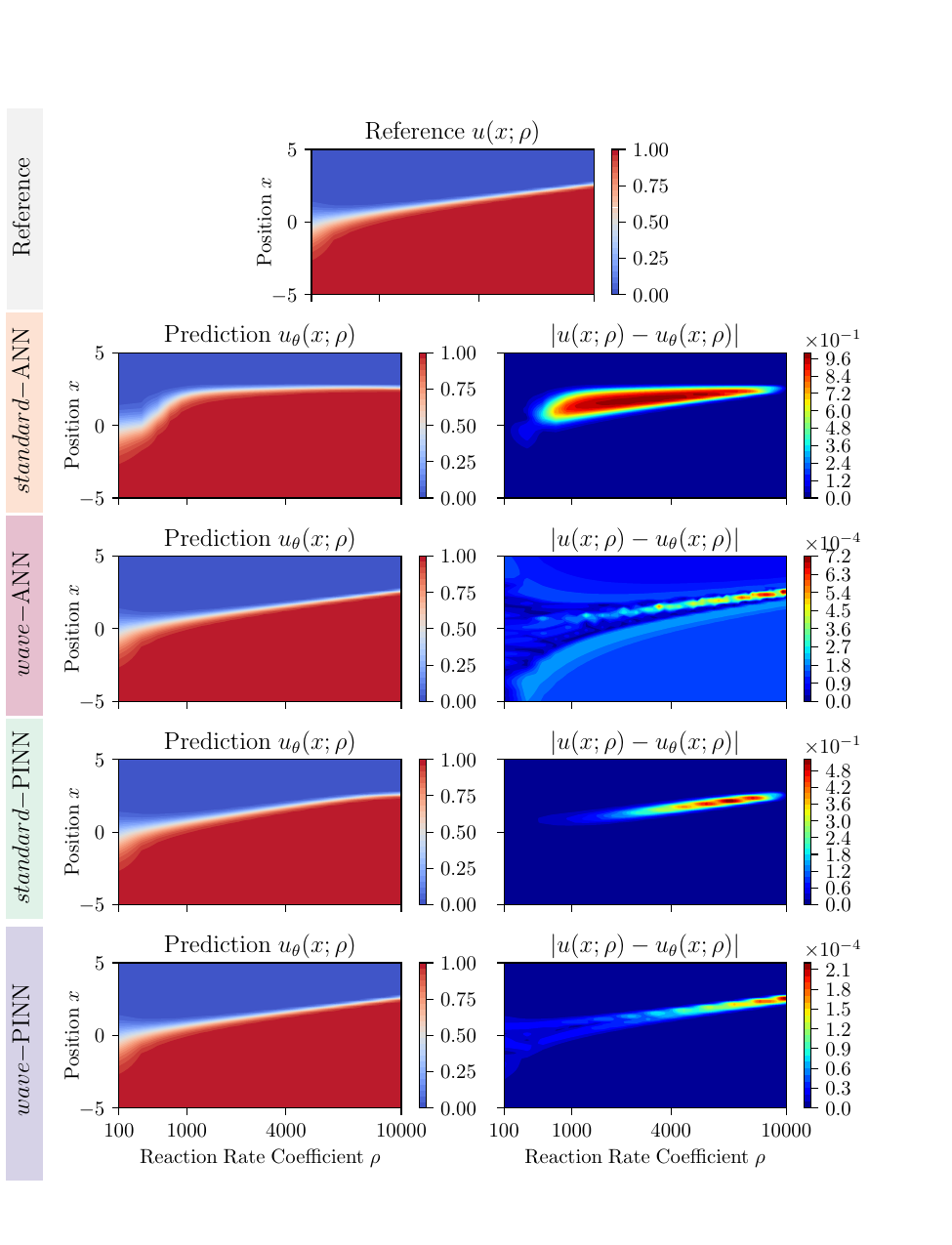}
    \caption{Prediction of the wave front $u(x)$ by a single model in the continuous $\rho$-space for the final time step $t=0.004$. For visualization purposes, the x-scale has been transformed using $\rho\leftarrow\sqrt{\rho}$.}
    \label{fig:continuous_rho_interpolation_2D}
\end{figure}

\para{Interpolation Performance.}
Having access to the solution function in the continuous form $u_\theta(x,t;\rho)$ allows to determine the $L_2$-error on the continuous $\rho$-domain where the modes were trained.
This consideration is depicted in Figure~\ref{fig:continuous_rho_interpolation}, which illustrates the error curve for the model trained in the domain $\rho\in[10^2, 10^4]$.
Additionally, the two-dimensional wave front $u(x;\rho)$ for the last time step $t=0.004$ is plotted and shown in Figure~\ref{fig:continuous_rho_interpolation_2D}.

Both figures demonstrate the outstanding performance of the \textit{wave}-ANN and \textit{wave}-PINN, with the latter clearly performing better in terms of the $L_2$-error (cf. Figure~\ref{fig:continuous_rho_interpolation}).
This is also reflected in Figure~\ref{fig:continuous_rho_interpolation_2D} where the \textit{wave}-PINN yields lower absolute errors along the continuous wave front $u(x;\rho)$.

\para{Extrapolation Performance.}
Both \textit{wave} models have indeed effectively learned the intrinsic structure of the parameterized solution function, as defined by Eq.~\eqref{eq:fishers_solution}.
This is evident in Figure~\ref{fig:continuous_rho_interpolation}, where both models exhibit excellent performance throughout the plotted $\rho$-domain, extending beyond the confines of the training domain.
Remarkably, even within the range of $\rho=10^5$, both models continue to provide strong approximations in the extrapolation regime, with only a slight degradation compared to the interpolation regime.

Comparing the two \textit{standard} models, it is evident that the PINN performs better than the data-driven ANN, though their performance is not competitive with the \textit{wave} models.

\section{Discussion \& Future Work}\label{sec:fishers_discussion}
The obtained results suggest that residual weighting schemes can overcome issues associated with sharp solutions as it is the case with steep propagating wave fronts in Fisher's equation. 
The results presented in Table~\ref{tab:discrete_rho} and shown in Figure~\ref{fig:fisher_wave_front} resonate with the general discussion of standard PINNs struggling with approximating sharp solutions.
Further modifications to the standard PINN framework are thus needed in order to overcome these difficulties.

The introduced residual weighting scheme is determined by the reaction term $F$, and reduces the weight of residuals in sharp regions with predominant reaction dynamics as evident from Figure~\ref{fig:fisher_wave_front}.
The use of residual-based weights in the optimization of PINNs has been previously tested and is often motivated by accelerated convergence~\cite{anagnostopoulos2023residual}, stiffness~\cite{mcclenny2020self}, and causality conformity~\cite{wang2024respecting}.
Furthermore, residual weights for sharp transitions in PDEs was proposed in~\cite{liu2024discontinuity}.
However, in~\cite{liu2024discontinuity} the weighting scheme relies on the compressible property of hyperbolic PDEs which, consequently, is not directly applicable to reaction-diffusion systems as they are parabolic.
Future work shall be dedicated to investigating the effectiveness of the proposed residual weighting scheme on more complex and general reaction mechanisms, e.g. of the form $\rho u(1-u^q)$ with $q>0$, as introduced in~\cite{kolmogorov1937study}.

The second investigation made in this work involved testing the \textit{wave} architecture, specifically designed to approximate wave-like solutions in the form of $u(\tilde z)$.
As evident from the obtained results, the additional introduction of the wave layer proved highly effective in approximating solutions that take this particular form.
The most remarkable difference between the \textit{standard} and \textit{wave} architecture is apparent in Table~\ref{tab:continuous_rho} and Figure~\ref{fig:continuous_rho_interpolation}.

One particularly interesting observation has been made in Table~\ref{tab:discrete_rho}, where the \textit{wave}-PINN exhibited superior performance compared to data-driven models explicitly trained on the analytical solution.
Originally, the aim was to establish a performance benchmark for this network size to evaluate how closely the performance of PINNs would align with it.
Surprisingly, the \textit{wave}-PINN surpassed the data-driven models.
Future investigations will aim to determine whether PINNs indeed provide a learning framework capable of surpassing purely data-driven models for specific learning tasks.

Finally, the \textit{generalizing} architecture has been tested, which, in contrast to the discrete-$\rho$ approximation, provides an approximation to an entire family of solutions to Fisher's equation in the form of $u(x,t;\rho)$. 
Extending the standard spatio-temporal input domain by an additional third dimension corresponding to the reaction rate coefficient $\rho$ offers a simple and effective approach to obtain multiple solutions at once by training only a single PINN instance.

However, approximating the solution function of parametric PDEs with PINNs is not novel. For instance, in~\cite{sun2020surrogate}, PINNs are extended to solve a parametric form of the incompressible Navier-Stokes equations by incorporating fluid properties as additional inputs to the network architecture. 
Furthermore, in~\cite{gao2021phygeonet}, physics-informed convolutional neural networks are presented for solving parametric PDEs on irregular domains. 
Recent advances in learning function-to-function maps have led to the development of operator learning techniques, with DeepONets serving as the primary workhorse~\cite{lu2021learning, wang2021learning}.
Additional literature on PINNs for parametric PDEs is provided by~\cite{hosseini2023single, mao2020physics}, and~\cite{penwarden2023metalearning}.

Still, several questions arise in the regard of learning the solution to parametric PDEs. 
For instance, what is the best approach to account for a particular parameterization? 
In the followed approach, the reaction rate coefficient $\rho$ appeared linearly in the governing differential equation, as shown in Eq.~\eqref{eq:fishers_equation}. 
Furthermore, the \textit{wave} architecture was used with a specific feature scaling (see Section~\ref{sec:experimental_setup_fisher}) that seemed highly suitable for the given problem. 
Specifically, by using $\rho_1 \leftarrow \sqrt{\rho}$ and $\rho_2 \leftarrow \rho$, a particular dependency was encoded into the model which was essential to train any of the \textit{wave} models, thereby reducing the sensitivity of the weights $\theta_j$ in the wave layer.
Importantly, this step was motivated by the commonly performed nondimensionalization step taking $\hat x\leftarrow \sqrt{\rho}x$ and $\hat t\leftarrow \rho t$.

Furthermore, having established a specific network architecture to account for the additional dimension, it would be interesting to know the best and most efficient approach to sampling the collocation points. 
Again, the linear dependence in the differential equations and the specific feature scaling seemed to align well with standard Latin hypercube sampling, which could explain the outstanding performance of the \textit{wave}-PINN.

Future work should focus on extending the insights gained from this study. 
In particular, it would be interesting to further expand the discussed approach to address more complex reaction-diffusion systems and explore alternative PINN methods for solving differential equations across continuous parameterizations.

\section{Conclusion}\label{sec:conclusion}
The rapid development of PINNs has made them powerful tools for approximating solutions to differential equations. 
In this study, PINNs were applied to solve Fisher’s equation with high reaction rate coefficients, where the solution features sharp traveling waves. 
To overcome the optimization difficulties associated with these sharp transitions, a novel residual weighting scheme was introduced into the optimization process. 
This scheme, based on the dynamics of the reaction term, facilitates the optimization in regions with steep gradients.
Additionally, a specialized network architecture was tested, specifically designed to model traveling waves. 
This architecture demonstrated significant improvements in accuracy compared to conventional network architectures.
Moreover, a PINN architecture incorporating the reaction rate coefficient as an input variable was explored, enabling the network to generalize across a family of solutions. 
This allows the model to approximate multiple sharp solutions to Fisher’s equation within a single instance.
The methods developed in this paper have potential applications in fields where sharp wave fronts and steep transitions are common, such as biological population dynamics, chemical reaction kinetics, and wave propagation in physical systems. 
The ability to efficiently model sharp solutions with large reaction rates offers new opportunities for simulating complex, real-world phenomena involving nonlinear differential equations.

\section*{Acknowledgements}
This work was supported by the the Austrian COMET — Competence Centers for Excellent Technologies — Programme of the Austrian Federal Ministry for Climate Action, Environment, Energy, Mobility, Innovation and Technology, the Austrian Federal Ministry for Digital and Economic Affairs, and the States of Styria, Upper Austria, Tyrol, and Vienna for the COMET Centers Know-Center and LEC EvoLET, respectively. The COMET Programme is managed by the Austrian Research Promotion Agency (FFG).

Additionally, the work by Franz M. Rohrhofer and Bernhard C. Geiger was supported by the European Union’s HORIZON Research and Innovation Programme under grant agreement No 101120657, project ENFIELD (European Lighthouse to Manifest Trustworthy and Green AI).

\bibliographystyle{unsrt}  
\bibliography{references}

\end{document}